\pdfoutput=1
\documentclass[11pt]{article}

\usepackage[final]{ACL2023}

\usepackage{times}
\usepackage{latexsym}

\usepackage[T1]{fontenc}

\usepackage[utf8]{inputenc}
\usepackage{microtype}

\usepackage{inconsolata}
\usepackage{CJKutf8}
\usepackage{amsmath}
\usepackage{amsfonts}
\usepackage{makecell}
\usepackage{multirow}
\usepackage{booktabs}
\usepackage{graphicx}
\usepackage{subfigure}
\usepackage{multicol}
\usepackage{microtype}
\usepackage{listings}
\usepackage{subcaption}
\usepackage{pifont}
\usepackage{xcolor}
\usepackage{colortbl}
\usepackage{soul}
\definecolor{'colour1'}{HTML}{E1F2EC}
\definecolor{'colour2'}{HTML}{FAFDFD}
\definecolor{'colour3'}{HTML}{FFF2E3}
\usepackage{listings} 

\newcommand{\illusion}[1]{%
  \begingroup
  \sethlcolor{yellow}%
  \textcolor{red}{\hl{#1}}%
  \endgroup
}

\definecolor{codegreen}{rgb}{0.7,0.9,0.7}
\definecolor{codegray}{rgb}{0.9,0.9,0.9}
\definecolor{codepurple}{rgb}{0.8,0.8,1}
\definecolor{backcolour}{rgb}{0.1,0.1,0.1}

\lstdefinestyle{mystyle}{
    backgroundcolor=\color{backcolour},   
    commentstyle=\color{codegreen},
    keywordstyle=\color{magenta},
    stringstyle=\color{codepurple},
    basicstyle=\ttfamily\footnotesize\color{codegray},
    breakatwhitespace=false,         
    breaklines=true,                 
    captionpos=b,                    
    keepspaces=true,                 
    numbers=none,  
    showspaces=false,                
    showstringspaces=false,
    showtabs=false,                  
    tabsize=2
}

\title{FinDABench: Benchmarking Financial Data Analysis Ability \\ of Large Language Models}

\author{
\textbf{Shu Liu}$^{1}$\hspace{0.5cm}
\textbf{Shangqing Zhao}$^{1}$\hspace{0.5cm}
\textbf{Chenghao Jia}$^{1}$\hspace{0.5cm}
\textbf{Xinlin Zhuang}$^{1}$\hspace{0.5cm}
\textbf{Zhaoguang Long}$^{1}$\hspace{0.5cm}
\\
\textbf{Jie Zhou}$^{1}\,^{2}$\hspace{0.5cm}
\textbf{Aimin Zhou}$^{1}\,^{2}$\hspace{0.5cm}
\textbf{Man Lan}$^{1}\,^{2}$\thanks{\hspace{0.15cm}Corresponding Author}\hspace{0.5cm}
\textbf{Qingquan Wu}$^{3}$\hspace{0.5cm}
\textbf{Chong Yang}$^{4}$\hspace{0.5cm}
\\
\normalsize $^1$ School of Computer Science and Technology, East China Normal University 
\\
\normalsize  $^2$Shanghai Institute of AI for Education, East China Normal University
\\
\normalsize $^3$PICC AMC \hspace{0.5cm}
\normalsize $^4$Bytedance.
\\
\texttt{\small {shuliu}@stu.ecnu.edu.cn, mlan@cs.ecnu.edu.cn}
}

\begin{document}
\maketitle
\begin{abstract}
Large Language Models (LLMs) have demonstrated impressive capabilities across a wide range of tasks. However, their proficiency and reliability in the specialized domain of financial data analysis, particularly focusing on data-driven thinking, remain uncertain. To bridge this gap, we introduce \texttt{FinDABench}, a comprehensive benchmark designed to evaluate the financial data analysis capabilities of LLMs within this context. \texttt{FinDABench} assesses LLMs across three dimensions: 1) \textbf{Foundational Ability}, evaluating the models' ability to perform financial numerical calculation and corporate sentiment risk assessment; 2) \textbf{Reasoning Ability}, determining the models' ability to quickly comprehend textual information and analyze abnormal financial reports; and 3) \textbf{Technical Skill}, examining the models' use of technical knowledge to address real-world data analysis challenges involving  analysis generation and charts visualization from multiple perspectives. We will release \texttt{FinDABench}, and the evaluation scripts at \url{https://github.com/cubenlp/BIBench}. \texttt{FinDABench} aims to provide a measure for in-depth analysis of LLM abilities and foster the advancement of LLMs in the field of financial data analysis.
\end{abstract}

\section{Introduction}

With the advance in pre-trained language models (PLMs) \cite{devlin2019BERTPretrainingDeep}, the Natural Language Processing (NLP) technology is evolving fast, so as its applications in financial  domains~\cite{zhang2023xuanyuan}. With the release of ChatGPT series~\cite{openai_chatgpt}, decoder-only Large Language Models (LLMs) like GPT-4~\cite{openai2023GPT4TechnicalReport} and LLaMA family~\cite{touvron2023llama, touvron2023llama2,llama3} have rapidly become a cornerstone of modern artificial intelligence, demonstrating remarkable versatility and power in NLP. The ability of LLMs to understand, generate and sometimes even reason with human language has led to transformative applications across numerous fields~\cite{huang2023CEvalMultiLevelMultiDiscipline, zhong2023AGIEvalHumanCentricBenchmark}. However, despite their broad capabilities, the performance of LLMs in specialized domains, particularly those requiring data-driven financial analytical skills, has not been thoroughly examined.

\begin{figure}[t!]
\centering
\includegraphics[width=0.5\textwidth]{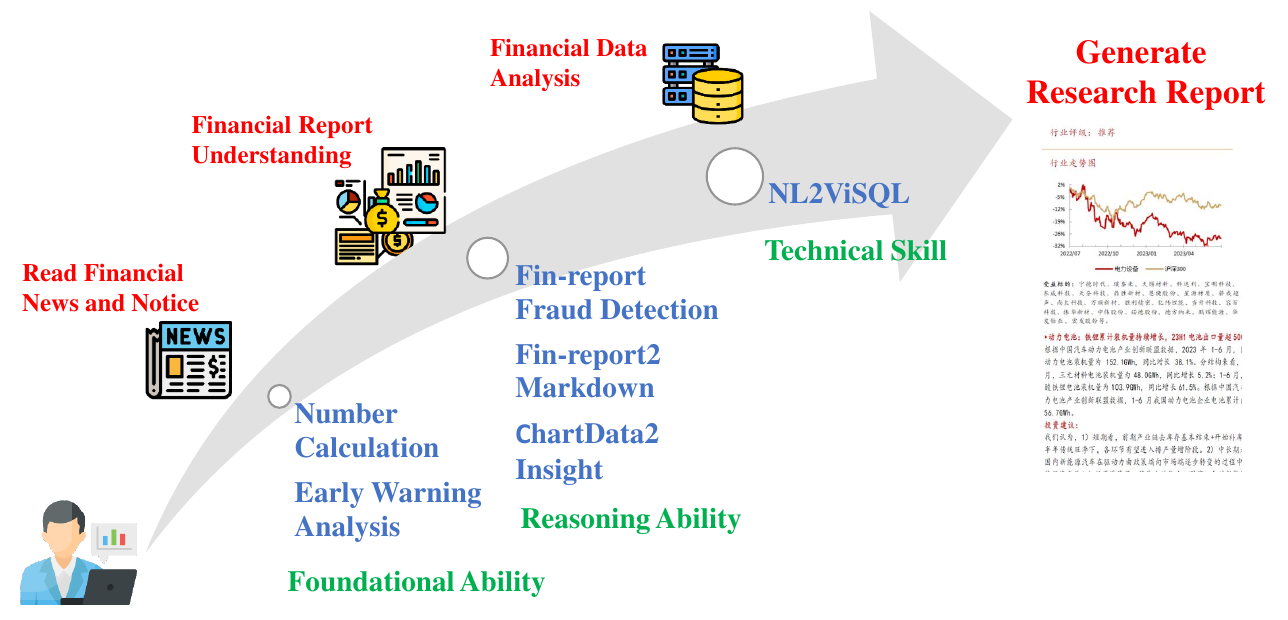}
\caption{
The \textcolor{red}{job skills} and their corresponding \textcolor{blue}{task names} required for financial analysts to manage daily work. Text highlighted in \textcolor{green}{green} denotes the standard capabilities of financial analysts.
} \label{FinancialDataAnalyst}
\end{figure}

Figure~\ref{FinancialDataAnalyst} illustrates the daily workflow of a financial analyst. \textbf{First}, analysts engage with news and company announcements, assess public sentiment, and calculate relevant metrics, tasks that represent the Foundational Ability. \textbf{Second}, they review corporate financial statements to extract data, evaluate anomalies, and formulate opinions—a demonstration of Reasoning Ability. \textbf{Lastly}, using data analysis techniques to derive insights and generate research reports exemplifies their Technical Skill.
This financial scenario stands in stark contrast to previous financial benchmarks like BBT-CFLEB\cite{lu2023bbt}, FinEval\cite{zhang2023FinEvalChineseFinancial}, and PIXIU~\cite{xie2023PIXIULargeLanguage}, which primarily focus on evaluating financial concepts through question-answering. Unlike these, financial data analysis demands the synthesis of information from diverse sources, formulation of pertinent questions, and application of advanced technical skills for in-depth data analysis and interpretation. These sophisticated requirements introduce unique challenges for LLMs, which have typically been assessed on more general language tasks.

\begin{figure}[t!]
\centering
\includegraphics[width=0.5\textwidth]{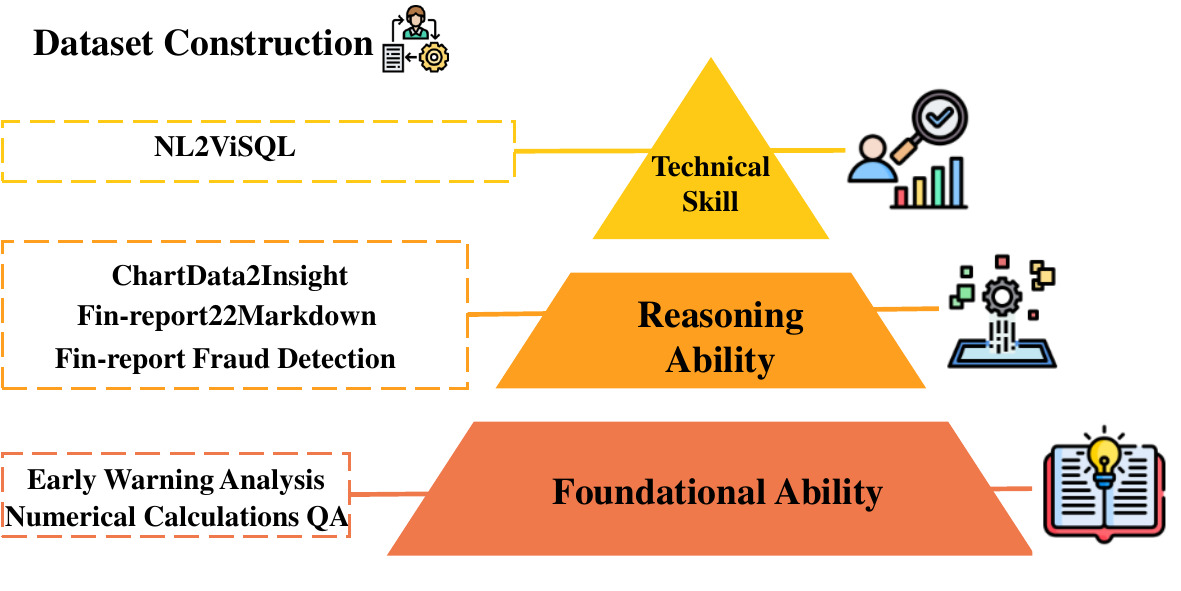}
\caption{\texttt{FinDABench} aims to provide a multi-faceted evaluation framework that mirrors the multifarious nature of financial data analysis tasks. 
} \label{framework}
\end{figure}

To address this challenge, we introduce \texttt{FinDABench}, a pioneering benchmark specifically designed to probe the depths of LLMs' data analysis capabilities within the financial domain. Inspired by \textit{Bloom's Taxonomy}~\cite{krathwohl2002revision} and \textit{Thinking, Fast and Slow}\cite{kahneman2011thinking, bengio2019system1to2}, which provide a widely recognized framework for categorizing tasks\cite{yu2023kola}, we developed a three-tiered framework to evaluate the financial data analysis capabilities of large models. The dataset framework diagram is shown in Figure~\ref{framework}. Organized around three core competencies, this taxonomy aims to deliver specific and insightful evaluation results, highlighting potential deficiencies in various aspects of financial data analysis.
\texttt{FinDABench} evaluates LLM skills that include domain-specific knowledge, including numerical reasoning(\textit{Numerical Calculations QA}) and corporate sentiment risk assessment(\textit{Early Warning Analysis}). The ability to extract relevant information from a variety of data sources is crucial, encompassing the analysis of chart (\textit{ChartData2Insight}) and the interpretation and anomaly detection of reports (\textit{Fin-report2Makrdown} and \textit{Fin-report fraud detection}). Furthermore, it is crucial to skillfully perform multi-perspective analysis and generate corresponding visualizations(\textit{NL2ViSQL}).

\texttt{FinDABench} comprises 6 sub-tasks, which fall under three categories of task types: classification, extraction, and generation. Together, these tasks constitute a comprehensive suite that rigorously tests the models across the spectrum of skills required in financial data analysis.
Our goal is to establish a standard for in-depth evaluation of LLMs in the context of finance and to catalyze further advancement in applying LLMs to data analysis. By doing so, we hope to bridge the gap between the capabilities of general-purpose LLMs and the specialized demands of financial data analysis, paving the way for more sophisticated and reliable AI tools in the realm of business and beyond.

Our contributions are summarized as follows:
\begin{itemize}
\item We introduce \texttt{FinDABench}, the first benchmark comprising six sub-tasks across three dimensions, designed to evaluate the financial data analysis capabilities of LLMs.

\item We systematically benchmark 41 popular LLMs’ financial data analysis capabilities for the first time. On top of their performance on \texttt{FinDABench}, we offer deep insights into the status quo of LLMs’ development and highlight the deficiencies that need improvements.

\item We evaluate the most recent methods on \texttt{FinDABench}. Our benchmark presents formidable challenges to existing methods. Notably, the SoTA GPT-4 achieves
merely a 32.37\% total result in zero-shot settings, while the performance of all other methods falls below 30\%.
\end{itemize}

\section{Related Work}

\begin{table*}[]\normalsize\centering
\resizebox{\linewidth}{!}{
\begin{tabular}{cccccc}
\hline
Benchmark                  & Data Source             & Evaluation angle                 & New Tasks                               & Dataset Systematics                     & \multicolumn{1}{l}{Open-ended}          \\ \hline
BBT-Fin~\cite{lu2023bbt}                    & Existing datasets       & Financial Knowledge              & \textcolor{red}{\ding{55}}              & \textcolor{red}{\ding{55}}              & \textcolor{green}{\ding{51}}          \\
FinEval~\cite{zhang2023FinEvalChineseFinancial}                    & Academic books          & Financial subject knowledge      & \textcolor{red}{\ding{55}}              & \textcolor{red}{\ding{55}}             & \textcolor{green}{\ding{51}}          \\
PIXUIU~\cite{xie2023PIXIULargeLanguage}                     & Existing datasets       & Financial Knowledge              & \textcolor{red}{\ding{55}}              & \textcolor{red}{\ding{55}}              & \textcolor{green}{\ding{51}}          \\
SuperCLUE-Fin~\cite{xu2024superclue}              & Exams \& Academic books & Financial Knowledge              & \textcolor{green}{\ding{51}}          & \textcolor{green}{\ding{51}}          & \textcolor{red}{\ding{55}}              \\ \hline
\textbf{FinDABench (ours)} & \textbf{Real Scenarios} & \textbf{Finanical data analysis} & \textcolor{green}{\ding{51}}  & \textcolor{green}{\ding{51}}  & \textcolor{green}{\ding{51}}  \\ \hline
\end{tabular}
}
\caption{
\textbf{Comparison of FinDABench with most recent financial benchmarks}: FinDABench is the \textbf{first} and the \textbf{only} benchmark that focuses on the financial data analysis domain. "New Tasks" means specific new evaluation tasks. "Dataset Systematics" means multi-level design evaluation system.} \label{datasetcompare}
\end{table*}

\subsection{Benchmarks for Large Language Models}

Concerning financial domain-specific competencies, 
\textbf{BBT-CFLEB}~\cite{lu2023bbt},
comprising six tasks, evaluates financial NLU and generation capabilities in dimensions including text summarization, question answering, classification, and relation extraction.
\textbf{FinEVal}~\cite{zhang2023FinEvalChineseFinancial} is a collection of high-quality multiple-choice questions spanning finance, economics, accounting, and certification domains. It consists of 4,661 questions across 34 different academic subjects.
\textbf{PIXIU}~\cite{luo2021synthesizing} aggregates 27 existing financial datasets, encompassing tasks such as semantic matching, sentiment analysis, entity recognition, and question answering, covering all aspects of financial natural language processing.
\textbf{SuperCLUE-Fin}~\cite{xu2024superclue} covers six real-world scenarios and 25 subtasks, evaluating models in financial contexts from two dimensions: foundational capabilities and application abilities.
These evaluations have notable limitations that prevent us from comprehensively assessing the financial data analysis capabilities of LLMs as exhibited in Table~\ref{datasetcompare}.

\begin{figure*}[t!]
\centering
\includegraphics[width=1\textwidth]{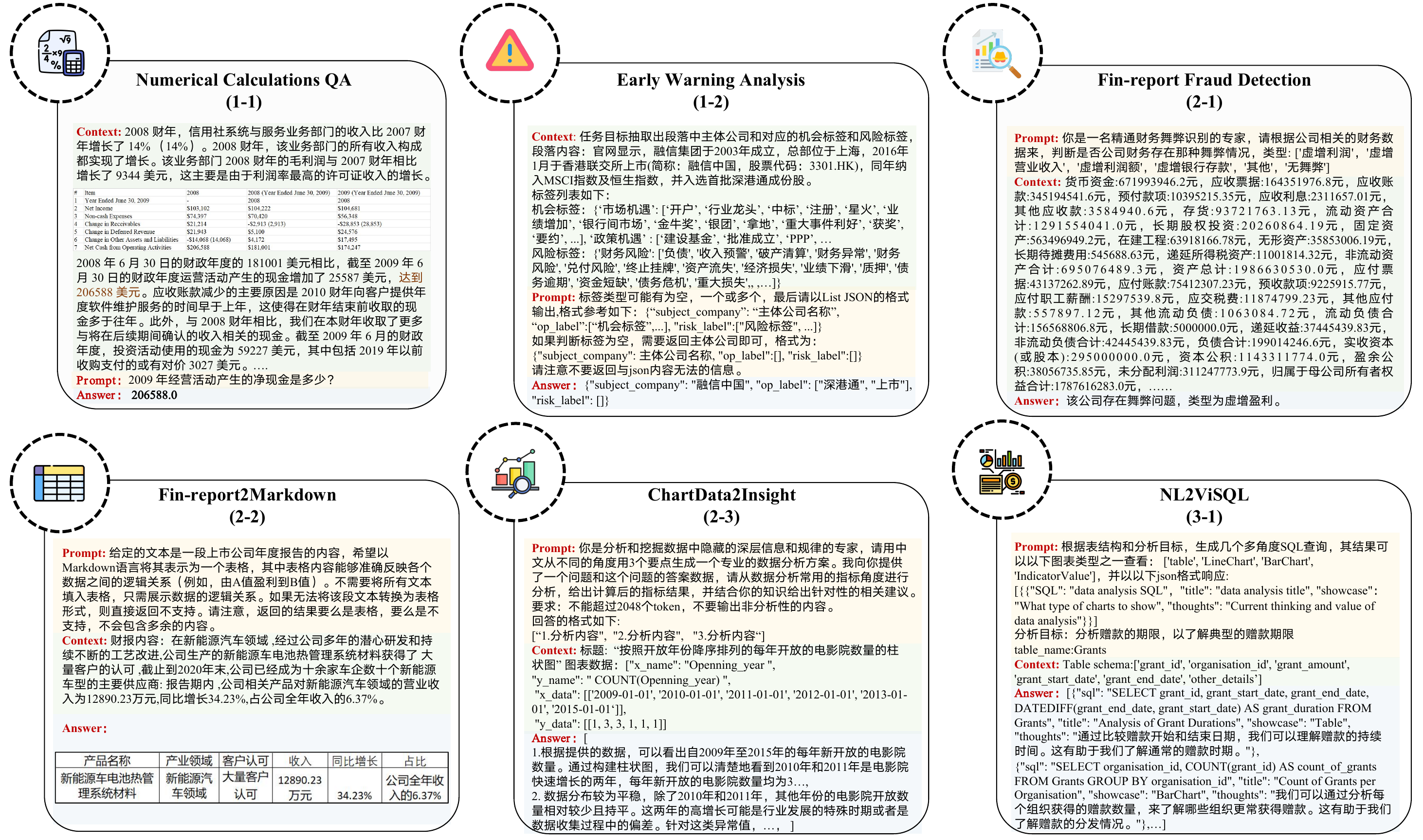}
\caption{
Data examples for the six sub-tasks of \texttt{FinDABench}, each including questions and answers with a unique identifier to facilitate differentiation. For the English version, please see the appendix\ref{sec:enversiondataexamples}.
} \label{datasample}
\end{figure*}

\subsection{Advancements in Large Language Models}  
In the realm of computational linguistics, there has been a profound and accelerating interest in Large Language Models (LLMs), which are trained on vast textual corpora. These models have shown remarkable ability in generating high-quality text across a spectrum of applications, both general and domain-specific \cite{llmsurvey1, llmsurvey2, llmsurvey3}. LLMs can be classified into two categories based on their availability: closed-source and open-source models.
Prominent examples of closed-source LLMs include the GPT-family,
Claude3,
Gemini2,
and ERNIEv4.0.
Nevertheless, there has been an increasing focus on open-source LLMs that provide comprehensive access to their model weights, facilitating deeper research exploration. A notable example is LLaMA-2 \cite{touvron2023llama2}, developed by Meta, which supports 20 languages and represents an evolution from its predecessor, LLaMA-1 \cite{touvron2023llama2}. The ChatGLM family \cite{du2022glm, zeng2022glm} offers multilingual models proficient in English, Chinese, and other languages. Additionally, Qwen \cite{bai2023QwenTechnicalReport} includes four model sizes, each with Base and Chat versions, the latter being optimized for human preferences.

\section{FinDA Benchmark}
We present \texttt{FinDABench}, the first benchmark comprising 2,400 instances specifically designed to evaluate the financial data analysis capabilities of LLMs and identify potential failure modes for each example. Subsequent sections will detail the guidelines for dataset construction based on task levels, describe \texttt{FinDABench}’s data and annotation structure, and present statistics of the dataset. Examples of these tasks are illustrated in Figure~\ref{datasample}.

\begin{figure*}[!t]
    \centering
    \subfigure[Interaction Rounds Data Distribution]{
    \includegraphics[width=3cm, height=2.8cm]{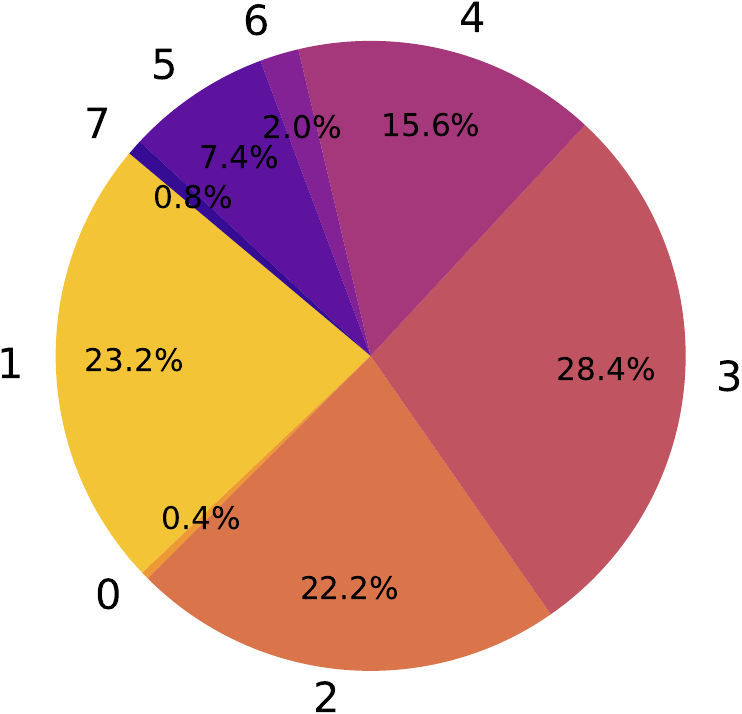}}
    \label{figvis:1-1}
    \hspace{5mm}
    \subfigure[Distribution of opportunity and risk labels]{
    \includegraphics[width=4cm, height=3cm]{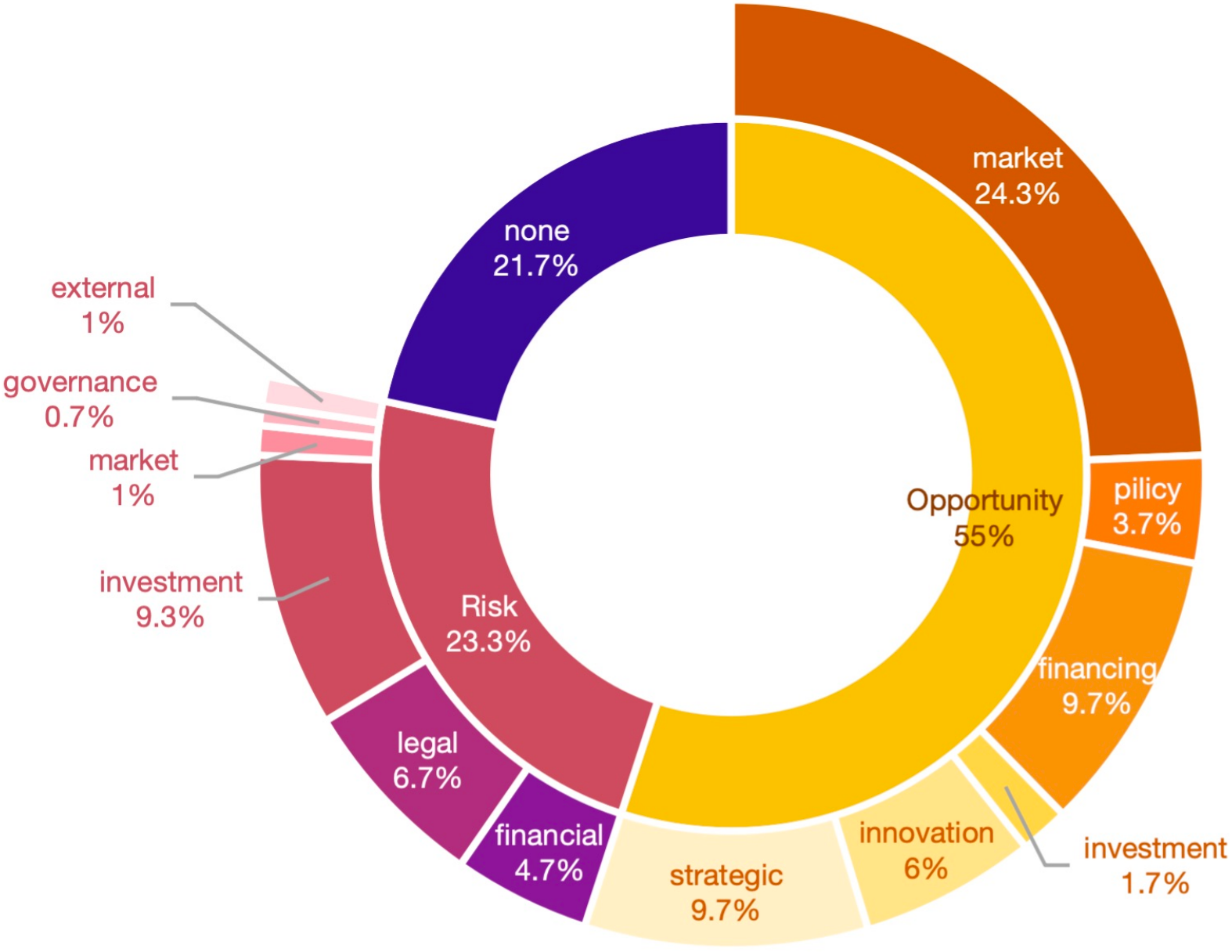}
    }
    \label{figvis:1-2}
    \hspace{5mm}
    \subfigure[Fraudulent labels distribution]{
    \includegraphics[width=4cm, height=2.8cm]{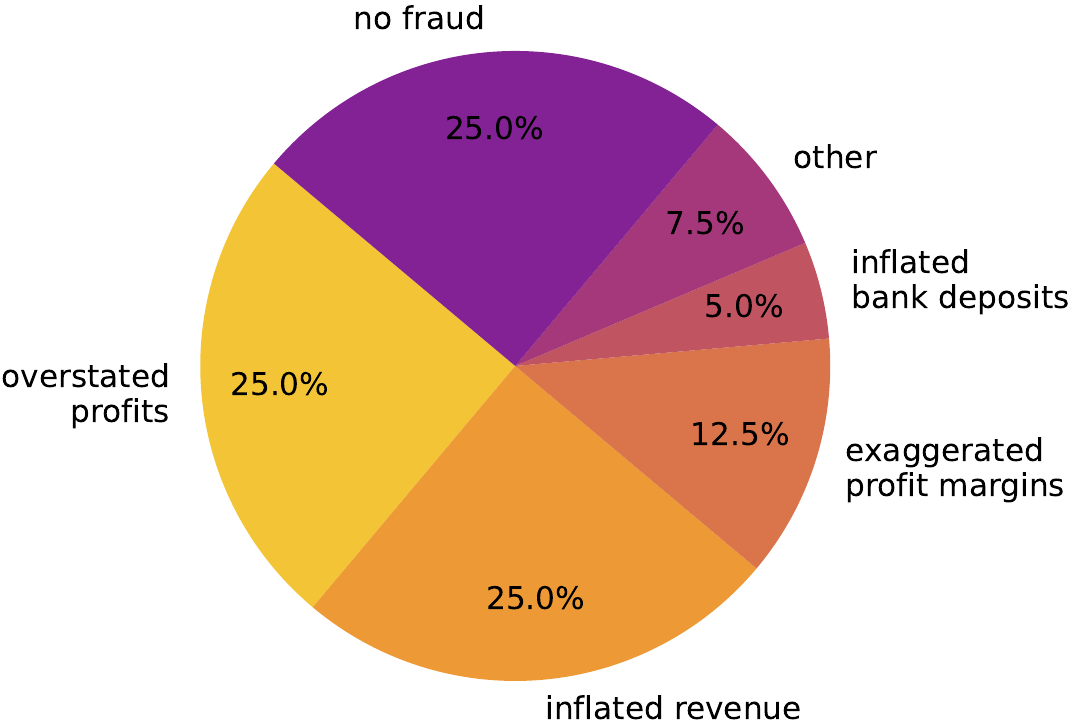}
    }
    \label{figvis:2-1}
    \subfigure[Length distribution]{
    \includegraphics[width=3.5cm, height=3cm]{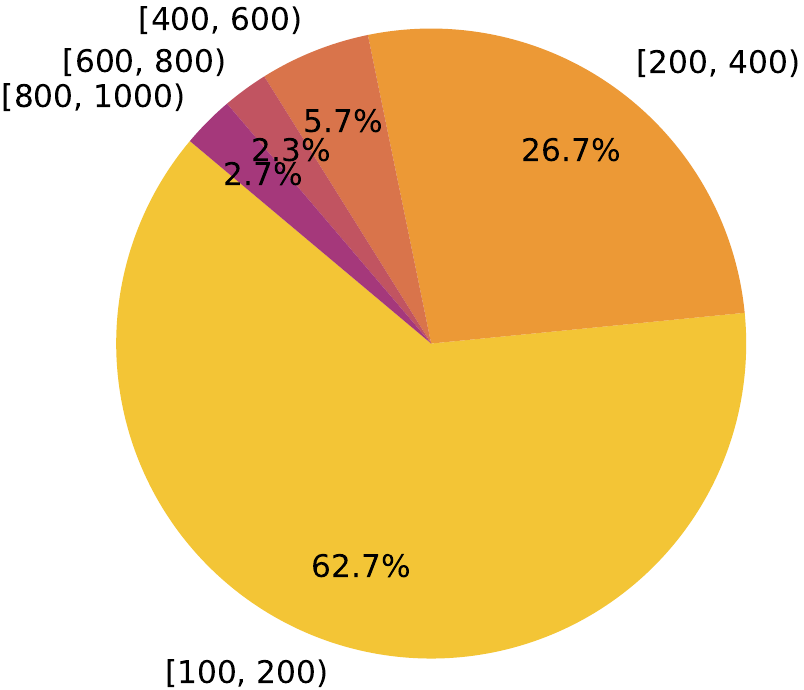}
    }
    \label{figvis:2-2}
    \hspace{12mm}
    \subfigure[Difficulty distribution]{
    \includegraphics[width=3cm, height=3cm]{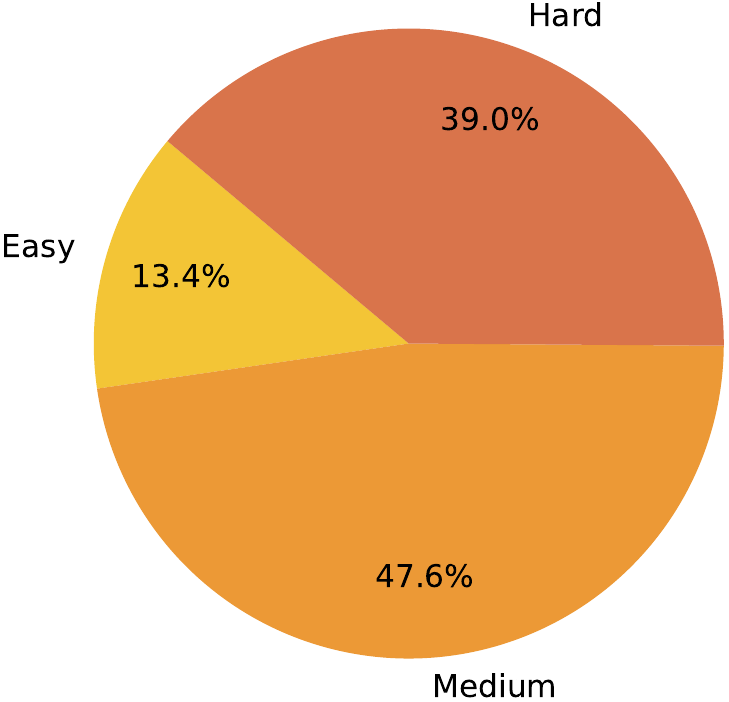}
    }
    \label{figvis:2-3}
    \hspace{12mm}
    \subfigure[Distinction distribution]{
    \includegraphics[width=3.8cm, height=2.8cm]{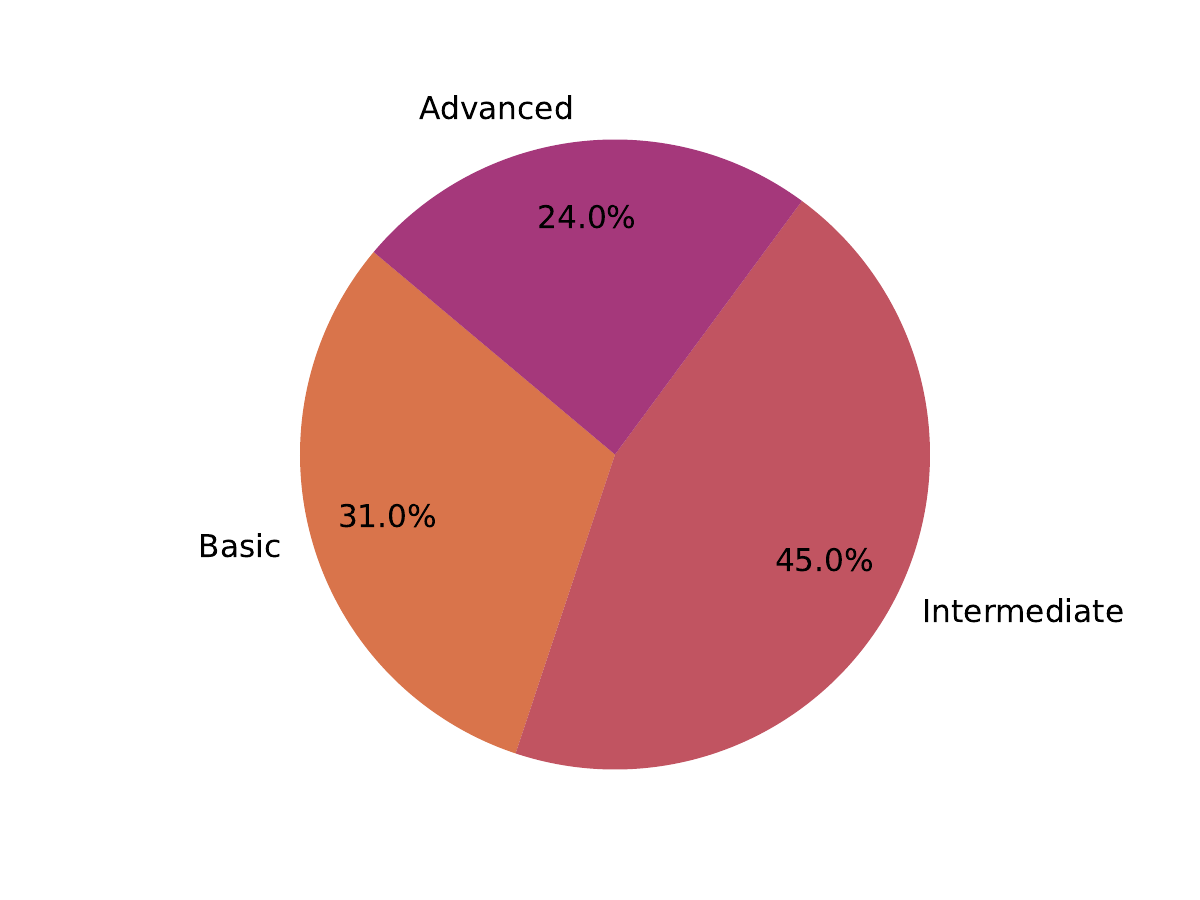}
    }
    \label{figvis:3-1}
    \caption{The statistical information for each sub-task of \texttt{FinDABench} is as follows: (a) represents Numberical Calculation QA, (b) represents Early Warning Analysis, (c) represents Fin-Report Fraud Detection, (d) represents Fin-Report2Markdown, (e) represents ChartData2Insight, and (f) represents NL2VisQL.}
    \label{fig:visualization}
\end{figure*}

\subsection{Foundational Ability}
The Foundational ability level measures essential skills for numerical computations and requires keen awareness of daily news that can impact financial markets. Professionals with this ability are equipped to interpret and respond to market fluctuations and news developments, providing the foundation for making timely and informed decisions.


\textbf{Numerical calculations QA} (1-1): \textit{Task definition: Numerical q\&a calculations based on text and tables from financial reports.}

Performing numerical calculations based on financial reports is a fundamental skill for financial analysts. We modified the ConvFinQA dataset~\cite{chen2022convfinqa} by first translating English financial reports and questions using GLM-4~\cite{du2022glm}. Specifically, we provided a translation prompt along with detailed requirements for the financial reports, which are outlined in Appendix\ref{sec:numericalcilationqa}.
As these reports contain both text and tables, and to prevent information loss during translation, we opted not to translate the table content, adhering instead to heuristic rules. After translation, manual checks ensured that the text conformed to the grammatical norms of the Chinese context. Additionally, we sampled 500 data entries based on the number of computational rounds, selecting samples with interaction counts ranging from zero to seven.

\textbf{Early Warning Analysis} (1-2): \textit{Task definition: extract the company entities from news, along with their associated opportunity and risk labels.}

Sentiment is one of the crucial indicators in financial data analysis for assessing the status of a company. Comprehensively evaluating a company's sentiment status, we have constructed a three-tier sentiment tagging system from a corporate perspective, set against the backdrop of the financial market and incorporating extensive industry expert experience. The primary labels are Opportunity labels (positive) and Risk labels (negative). Opportunity labels include secondary labels that represent potential opportunities such as market, policy, financing, investment, innovation, and strategic opportunities, with a total of 76 tertiary sub-labels. Risk labels encompass secondary labels for potential challenges including financial, legal, investment, market, governance, and external risks, with a total of 69 tertiary sub-labels. A detailed description of the labeling system is in Appendix\ref{sec:earlywarning}.

We scraped 600 company news articles from financial news websites and used regular expressions to extract the news summaries. After filtering out duplicates and irrelevant content, we retained 300 news summaries. Initially, we used sentiment keywords for rough labeling and then conducted a manual review to ensure the accuracy of the labels.

\subsection{Reasoning Ability}
The reasoning ability level demands a deep understanding of financial reports, surpassing basic data comprehension. It involves discerning potential fraud in financial statements and conducting in-depth analyses of chart data. Professionals with these skills can interpret explicit content and critically assess an organization's financial health and integrity, thus offering valuable insights.

\begin{table}[t!]
\resizebox{\linewidth}{!}{
\begin{tabular}{@{}cccccc@{}}
\toprule
\textbf{Cognitive Level}                                                                  & \textbf{ID} & \textbf{Task}              & \textbf{Data size} & \textbf{Metric} & \textbf{Type}  \\ \midrule
\multirow{2}{*}{\textbf{\begin{tabular}[c]{@{}c@{}}Foundational \\ Ability\end{tabular}}} & 1-1         & Numerical Reasoning QA     & 500                & Accuracy        & Generation     \\
& 1-2   & Early Warning Analysis     & 300  & F1     & Extraction     \\ \midrule
\multirow{3}{*}{\textbf{\begin{tabular}[c]{@{}c@{}}Reasoning\\ Ability\end{tabular}}}     & 2-1         & Fin-report fraud detection & 400                & F1              & Classification \\
 & 2-2   & Fin-report2Markdown& 300    & Rouge     & Generation     \\
& 2-3    & Data2Insight  & 500  & Rouge         & Generation     \\ \midrule
\textbf{\begin{tabular}[c]{@{}c@{}}Technical\\ Skill\end{tabular}}                        & 3-1         & NL2ViSQL                   & 400                & EM              & Generation     \\ \bottomrule
\end{tabular}
}
\caption{Basic information for \texttt{FinDABench}.} 
\label{findadataset}
\end{table}

\textbf{Fin-report Fraud Detection} (2-1): \textit{Task definition: Given financial report data, determine whether the financial statements are fraudulent.}

Determining whether a company's financial data involves fraud is foundational for subsequent analytical research. Based on the Securities Regulatory Commission's penalty announcements~\footnote{\url{http://www.csrc.gov.cn/csrc/xwfb/index.shtml}} and the expertise of financial experts, we categorize financial fraud into six types: \textbf{overstated profits}, \textbf{inflated revenue}, \textbf{exaggerated profit margins}, \textbf{inflated bank deposits}, \textbf{other}, and \textbf{no fraud}. We obtained the names of companies involved in financial fraud from the Commission's penalty announcements and downloaded the corresponding financial reports. We then extracted the key accounting data from the financial statement tables in these reports and performed manual annotations, ultimately generating 400 benchmark data entries.

\textbf{Fin-report2Markdown} (2-2): \textit{Task definition: Convert essential unstructured information from financial reports into a Markdown table.}

Extracting and converting unstructured data into tabular format showcases a financial analyst's analytical skills. We downloaded 300 PDF financial reports from the Shanghai Stock Exchange~\footnote{\url{https://www.sse.com.cn}}. Using the PDF parsing tool pdfumber, we extracted unstructured content based on chapter structure, ensuring paragraph integrity. Based on the expertise of financial professionals, Section 3 of these reports (Company Overview/Management Discussion and Analysis) often contains crucial data; thus, we selected this section as the unstructured data for conversion. We utilized GPT-4 for data annotation, providing it with specific prompt and detailed requirements for financial reports, as detailed in Appendix~\ref{sec:finreport2markdown}. Finally, the data underwent manual review and correction to ensure accuracy.

\textbf{ChartData2Insight} (2-3): \textit{Task definition: Generate data analysis suggestions and insights from the given chart data.} 

Generating viewpoints from chart data showcases the data reasoning skills of financial analysts. We selected 500 finance-related data entries from nvBench's~\cite{luo2021nvbench} charts, categorized by difficulty into Easy, Medium, and Hard levels. During the annotation process, we first translated queries in the data into Chinese, treating these queries as captions for the charts. We then fed X-axis and Y-axis data, along with the captions, into GPT-4. In particular, we provided it with prompt and specific requirements for chart data, as detailed in the Appendix\ref{sec:chartdatainsight}. Finally, the insights were reviewed by two senior financial data analysts.

\subsection{Technical Skill}
The Technical Skill demands that large language models embrace data-centric thinking and master external tools like SQL for sophisticated data analyses. This proficiency enables analysts to devise diverse analytical strategies, select optimal visualization types, and generate executable queries. With these skills, analysts can clearly translate complex datasets into actionable insights, boosting data interpretation and utility.

\textbf{NL2ViSQL} (3-1): 
\textit{Task definition: Generate SQL analysis statements from given questions and table structures, considering multiple perspectives.} 

Generating multi-perspective data analyses and visualizations from databases is an advanced capability for financial analysts. Using the single-table structure from Spider~\cite{yu2019spider}, we first employed few-shot learning with GPT-4 to align data analysis goals closely with real-world scenarios for each single-table; detailed instructions for this approach are presented in Appendix~\ref{sec:nl2visql}. We defined four visualization chart types: Table, LineChart, BarChart, and IndicatorValue, and required annotators to justify their SQL queries. Two senior financial analysts crafted multi-perspective SQL queries and selected appropriate visualization types based on the table structure and objectives. Additionally, we categorized these tasks by difficulty levels: Basic, Intermediate, and Advanced.

\begin{figure*}[ht!]
\centering
\includegraphics[width=1\textwidth]{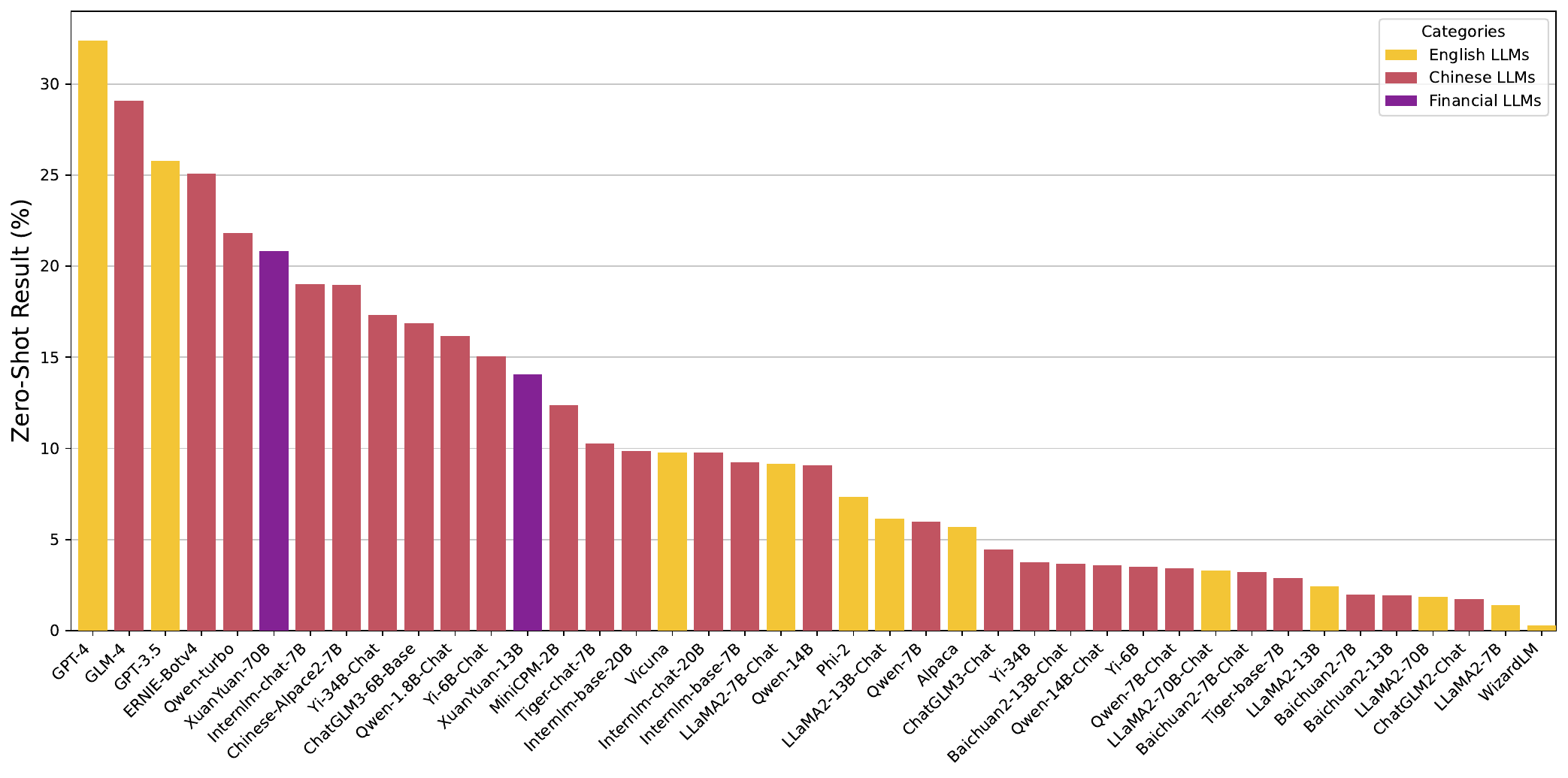}
\caption{Average performance (zero-shot) of 41 LLMs evaluated on \texttt{FinDABench}
} \label{barchart}
\end{figure*}

\section{Experiment}

\subsection{Dataset Statistics}
Table~\ref{findadataset} displays the count, evaluation metrics, and types for each sub-task. The Foundational Ability comprises 800 data entries, the Reasoning Ability includes 1300 entries, and the Technical Skill has 400 entries, along with the task types and evaluation metrics for each sub-task. Details of the sub-task data distribution are shown in Figure~\ref{fig:visualization}. Figure~\ref{fig:visualization} (a) and (b) describe the data distribution for Foundational Ability, with (b) showing that opportunity labels account for 55\% and risk labels for 23.3\%. The other pie charts follow similarly.

\subsection{Evaluation Metrics}
We defined 4 different metrics in total to measure different types of tasks:

\textbf{Accuracy}: Accuracy is a binary score that performs an exact match between the model prediction and the gold answer. This applies to single-label tasks, including tasks 1-1.
\textbf{F1}: When there are multiple output labels, the F1 score measures the harmonic mean of the precision and recall. This applies to all multi-label classification tasks, including tasks 2-1 and 2-2. 
\textbf{EM}(Exact Set Match): This metric compares the generated View SQL with the Gold SQL to ensure consistency in the structure's elements. It is applied to task 3-1.
\textbf{Rouge}: For other generation tasks 2-2 and 2-3, we use the Rouge(Rouge-1, Rouge-2, Rouge-L) score to evaluate them. Rouge-L, commonly used in the evaluation of generation tasks, automatically identifies the longest co-occurring n-gram sequences to compare the structural similarity of extracted answers with standard answers~\cite{lin2004rouge}. The formula is expressed as follows.
\begin{equation}
\text{$F_{lcs}$} = \frac{(1 + \beta^2) \cdot \text{Precision} \cdot \text{Recall}}{\text{Recall} + \beta^2 \cdot \text{Precision}} \\
\end{equation}

\subsection{Evaluated Models}
We evaluate a wide spectrum of large language models of various sizes, grouping them into three
major categories based on their pre-training and fine-tuning domains: English LLMs, Chinese
LLMs and Financial LLMs. We provide a short review of them in the following
section. The detailed model list is shown in Appendix Table~\ref{LLMTest}.

\textbf{English LLMs:} We consider 9 open-source English models: 
LLaMA-2-7B / 13B / 70B, LLaMA-2-Chat-7B / 13B / 70B, Alpacav1.0-
7B, Vicuna-v1.3-7B / 13B / 33B, WizardLM-7B. In addition, two commercial
models, GPT-3.5-turbo-0613 and GPT-4-0613, are included.

\begin{table*}[]
\resizebox{\linewidth}{!}{
\begin{tabular}{@{}cccccccccccccccc@{}}
\toprule
\multirow{2}{*}{Type}                                 & \multirow{2}{*}{Model}       & 1-1  & \multicolumn{3}{c}{1-2}   & \multicolumn{3}{c}{2-1}   & \multicolumn{3}{c}{2-2} & \multicolumn{3}{c}{2-3} & 3-1  \\ \cmidrule(l){3-16} 
  &    & ACC  & Precision & Recall & F1   & Precision & Recall & F1   & R-1    & R-2    & R-L   & R-1    & R-2    & R-L   & EM   \\ \midrule
\multicolumn{1}{c|}{\multirow{6}{*}{English Model}}   & LLaMA2-7B-Chat$_{0-shot}$ & 0.70 & 0.06  & 0.30   & 0.10 & 36.00      & 12.00   & 18.00 & 24.67   & 13.23   & 18.39  & 3.67   & 0.23  & 0.50  & 5.72 \\
\multicolumn{1}{c|}{}                                 & LLaMA2-7B-Chat$_{3-shot}$  & 0.92 & 0.23  & 0.50   & 0.32 & 28.43      & 15.70   & 20.23 & 28.72   & 14.36   & 23.58  & 7.63   & 0.78  & 1.23  & 7.21 
\\
\multicolumn{1}{c|}{}                                 & GPT-3.5$_{0-shot}$  & 0.81 & 25.47      & 23.23   & 24.30 & 21.96 & 32.76 & 26.30 & 37.25   & 14.87   & 24.17  & 23.73   & 12.84   & 10.95  & 9.89 \\
\multicolumn{1}{c|}{}                                 & GPT-3.5$_{3-shot}$  & 2.93 & 20.57      & \underline{34.79}   & 25.86 & 38.70 & \underline{42.36} & \underline{40.45} & 42.32   & 16.53   & 28.32  & \underline{29.37}   & \underline{14.67}   & 16.37  & \underline{11.78} \\
\multicolumn{1}{c|}{}                                 & GPT-4$_{0-shot}$   & 10.30 & 66.26   & 78.27   & 71.77 & 52.37   & \textbf{64.08}   & 57.64 & 42.36  & 18.27   & 29.51  & 20.89   & 9.27 & 10.27  & 10.21 \\
\multicolumn{1}{c|}{}                                 & GPT-4$_{3-shot}$   & \underline{15.45} & \textbf{84.13}   & \textbf{80.56}   & \textbf{82.31} & \textbf{72.59}   & 59.19   & \textbf{65.21} & \textbf{48.67}  & \underline{19.34}   & 33.27  & 23.26   & 10.81 & 13.45  & 11.01 
\\ \midrule
\multicolumn{1}{c|}{\multirow{10}{*}{Chinese Model}}  & GLM-4$_{0-shot}$  & 3.64 & \underline{37.46} & 15.60   & 22.03 & 18.91  & 12.62  & 15.14 & 39.26   & 15.28   & 25.75  & 25.87   & 12.36   & 14.79  & 10.58 \\
\multicolumn{1}{c|}{}                                 & GLM-4$_{3-shot}$  & 9.45 & 29.87 & 27.56   & 28.67 & \underline{42.29}  & 22.46  & 29.34 & 41.37   & 17.36   & 26.35  & \textbf{29.76}   & \textbf{15.35}   & \textbf{17.84}  & \textbf{12.58} \\
\multicolumn{1}{c|}{}                                 & ERNIEv4$_{0-shot}$  & 2.99 & 15.63  & 24.98 & 19.23 & 14.27 & 10.54 & 12.13 & 38.26 & 14.59 & 24.32  & 23.67   & 11.31   & 13.56  & 9.62 \\
\multicolumn{1}{c|}{}                                 & ERNIEv4$_{3-shot}$  & 7.26 & 18.32  & 32.92 & 23.54 & 29.32 & 21.15 & 24.58 & 39.41 & 15.21 & 25.75  & 25.83   & 13.42   & \underline{16.93}  & 10.32 \\
\multicolumn{1}{c|}{}                                 & Qwen-turbo$_{0-shot}$  & 8.49 & 18.34  & 13.57 & 15.60 & 21.79  & 20.45  & 21.10 & 44.20 & 17.86 & 32.27  & 10.23 & 3.23 & 1.62  & 5.72 \\
\multicolumn{1}{c|}{}                                 & Qwen-turbo$_{3-shot}$  & 12.32 & 24.90  & 15.78 & 19.32 & 27.36  & 23.56  & 25.32 & 46.31 & 18.68 & 35.57  & 13.42 & 5.76 & 8.89  & 8.63 \\
\multicolumn{1}{c|}{}                                 & Internlm-chat-7B$_{0-shot}$ & 1.66 & 30.72 & 27.42  & 28.98 & 17.96 & 23.73 & 20.45 & 42.12  & 15.96 & 29.48  & 8.23 & 2.12   & 1.25  & 5.34 
\\
\multicolumn{1}{c|}{}                                 & Internlm-chat-7B$_{3-shot}$ & 5.27 & 32.98 & 29.67  & \underline{31.24} & 17.30 & 20.22 & 18.65 & 43.46  & 18.26 & 31.27  & 9.62 & 4.41   & 3.21  & 7.52 
\\
\multicolumn{1}{c|}{}                                 & Yi-34-Chat$_{0-shot}$  & 7.26 & 13.08  & 15.31 & 14.11 & 8.23  & 6.93   & 7.53 & 42.08   & 15.58  & 29.41  & 8.02 & 2.01   & 1.11  & 3.45 \\
\multicolumn{1}{c|}{}                                 & Yi-34-Chat$_{3-shot}$  & 9.23 & 15.03  & 18.23 & 16.48 & 12.58 & 8.82 & 10.37 & 42.35 & 15.87 & 30.23  & 10.57  & 3.26 & 5.87  & 5.89 \\ \midrule
\multicolumn{1}{c|}{\multirow{4}{*}{Financial Model}} & XuanYuan-13B$_{0-shot}$  & 8.24 & 18.97 & 11.59 & 14.39 & 10.48 & 20.94  & 13.97 & 38.75 & 14.39  & 25.17  & 6.29   & 1.12   & 0.85  & 14.08 \\
\multicolumn{1}{c|}{}                                 & XuanYuan-13B$_{3-shot}$  & 10.29 & 20.35 & 16.49 & 18.22 & 18.13 & 16.72 & 17.40 & 38.82 & 14.75  & 26.82  & 6.71   & 2.37   & 2.38 & 2.56 \\
\multicolumn{1}{c|}{}                                 & XuanYuan-70B$_{0-shot}$  & 11.23 & 28.96  & 18.26  & 22.40 & 27.93  & 17.25 & 21.33 & 47.21 & 19.32  & 36.28 & 9.68  & 2.31  & 5.87  & 4.30 \\
\multicolumn{1}{c|}{}                                 & XuanYuan-70B$_{3-shot}$  & \textbf{18.3} & 30.42  & 23.71  & 26.65 & 18.90  & 23.70  & 21.03  & \underline{48.52}  & \textbf{20.76}  & \textbf{38.67}  & 13.02 & 4.96  & 8.67  & 8.72 \\ \bottomrule
\end{tabular}
}
\caption{Fine-grained results of \texttt{FinDABench}: Performance of various LLMs on the detailed sub-tasks in zero-shot and few-shot scenarios. The best results are highlighted in \textbf{bold}, and the second-best results are \underline{underlined}.}
\label{finegrainedresult}
\end{table*}

\textbf{Chinese LLMs:} A number of Chinese LLMs have been proposed to enhance Chinese comprehension. They typically perform better than English models on Chinese NLP tasks. We
include 24 open-sourced, Chinese LLMs in our evaluation: Yi-Base 6B/34B, Yi-Chat 6B/34B, InternLM-Base 7B/20B, InternLM-Chat 7B/20B, Qwen-Base 7B/14B, Qwen-Chat 7B/14B, Baichuan2-Base 7B/13B, Baichuan2-Chat 7B/13B, TigerBot-Base-7B, TigerBot-Chat-7B, Chinese-Alpace2-7B, ChatGLM-6B, ChatGLM2-6B, ChatGLM3-Base-6B, ChatGLM3-6B. Moreover, three commercial models, Qwen-turbo (\begin{CJK*}{UTF8}{gbsn}通义千问\end{CJK*}), ERNIEv4.0 (\begin{CJK*}{UTF8}{gbsn}文心一言\end{CJK*}) and GLM-4 (\begin{CJK*}{UTF8}{gbsn}智谱清言\end{CJK*}), are included.

\textbf{Financial LLMs:} XuanYuan-Chat based on LLaMA2-13B/70B, on a Chinese financial corpus to enhance their understanding of Chinese finances. Through \texttt{FinDABench}, we can rigorously assess their advancements and identify limitations compared to general-purpose LLMs, offering insights into both general and financial applications.

\subsection{Experiment Setting}
In the commercial models, we set the temperature to 0.7 and top p to 1. In other chat models, we tailor the prompt by using specific prefixes and suffixes for each model. Greedy decoding is performed during the generation process for all open-source models. We set the token length limit to 2400. Right truncation is performed for input prompts exceeding the length limitation. We evaluate all models in zero-shot and few-shot settings.

\begin{table}[t!]
\resizebox{\linewidth}{!}{
\begin{tabular}{@{}c|c|c|c@{}}
\toprule
Model         & Base Score & SFT Score & Diff Score \\ \midrule
Yi-6B      & 3.42     & 10.23   & 6.81     \\ \midrule
Tiger-7B      & 4.31     & 6.87   & 2.56     \\ \midrule
Internlm-7B   & 10.23     & 21.16   & 10.93     \\ \midrule
Baichuan2-13B & 1.92     & 3.67    & 1.91     \\ \midrule
Yi-34B        & 8.73     & 12.34   & 3.61    \\ \bottomrule
\end{tabular}
}
\caption{avg Performance score comparison of open source LLMs before and after SFT.} \label{scorecompare}
\end{table}

\subsection{Main Results}
Figure~\ref{barchart} displays the overall zero-shot performance of each model. GPT-4 and GLM-4 are significantly ahead in the benchmarks, vastly outperforming all other models. With the same model size, LLMs that underwent Supervised Fine-Tuning (SFT) in Chinese outshine both the base Chinese LLMs and English SFT LLMs, demonstrating the effectiveness of fine-tuning on Chinese data. Furthermore, recent smaller models, like MiniCPM-2B~\cite{hu2024minicpm}, also exceed the performance of many larger LLMs, indicating that the relationship between an LLM's capabilities and its size is not linear. Lastly, Financial LLMs surpass general LLMs, suggesting that domain-specific fine-tuning can enhance a model's domain capabilities.

In Table~\ref{finegrainedresult}, we display the fine-grained scores of different model configurations across all tasks. We made several observations. \textbf{First}, there is substantial variation in the distribution of scores across tasks. The best-performing model, such as GPT-4, can score over 60 in tasks 1-2 and 2-1 but does not exceed 30 in tasks 3-1 and 2-3. This demonstrates that our benchmark effectively assesses model capabilities in various aspects. \textbf{Second}, it is evident that scores under few-shot conditions are consistently higher than those under zero-shot across all model types. \textbf{Third}, it is promising that most LLMs exhibit some capability in handling financial data analysis tasks, yet there is still considerable room for improvement. Even the top-performing model, GPT-4, achieves only an average score of 32.37\% in zero-shot and 39.38\% in few-shot, highlighting the need for further efforts in the future.

\begin{figure}[t!]
\centering
\includegraphics[width=0.5\textwidth]{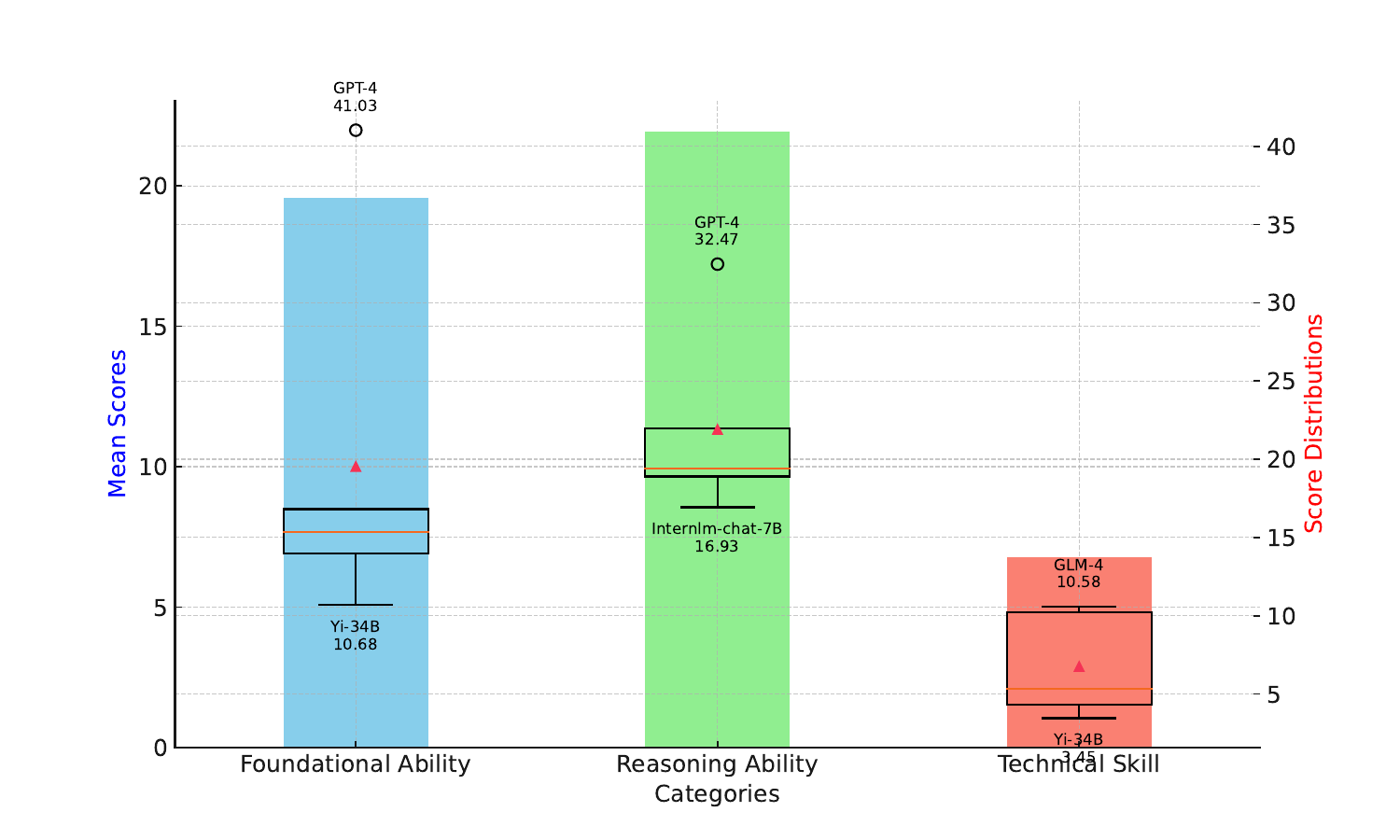}
\caption{
Display the \textcolor{blue}{average scores} and variance for the GPT-4, GLM4, XuanYuan-70B, Yi-34B, and Internlm-chat-7B models across three dimensions, showing only the \textcolor{red}{highest and lowest scores} for each model.
} \label{Histogram}
\end{figure}

\begin{figure}[t!]
\centering
\includegraphics[width=0.5\textwidth]{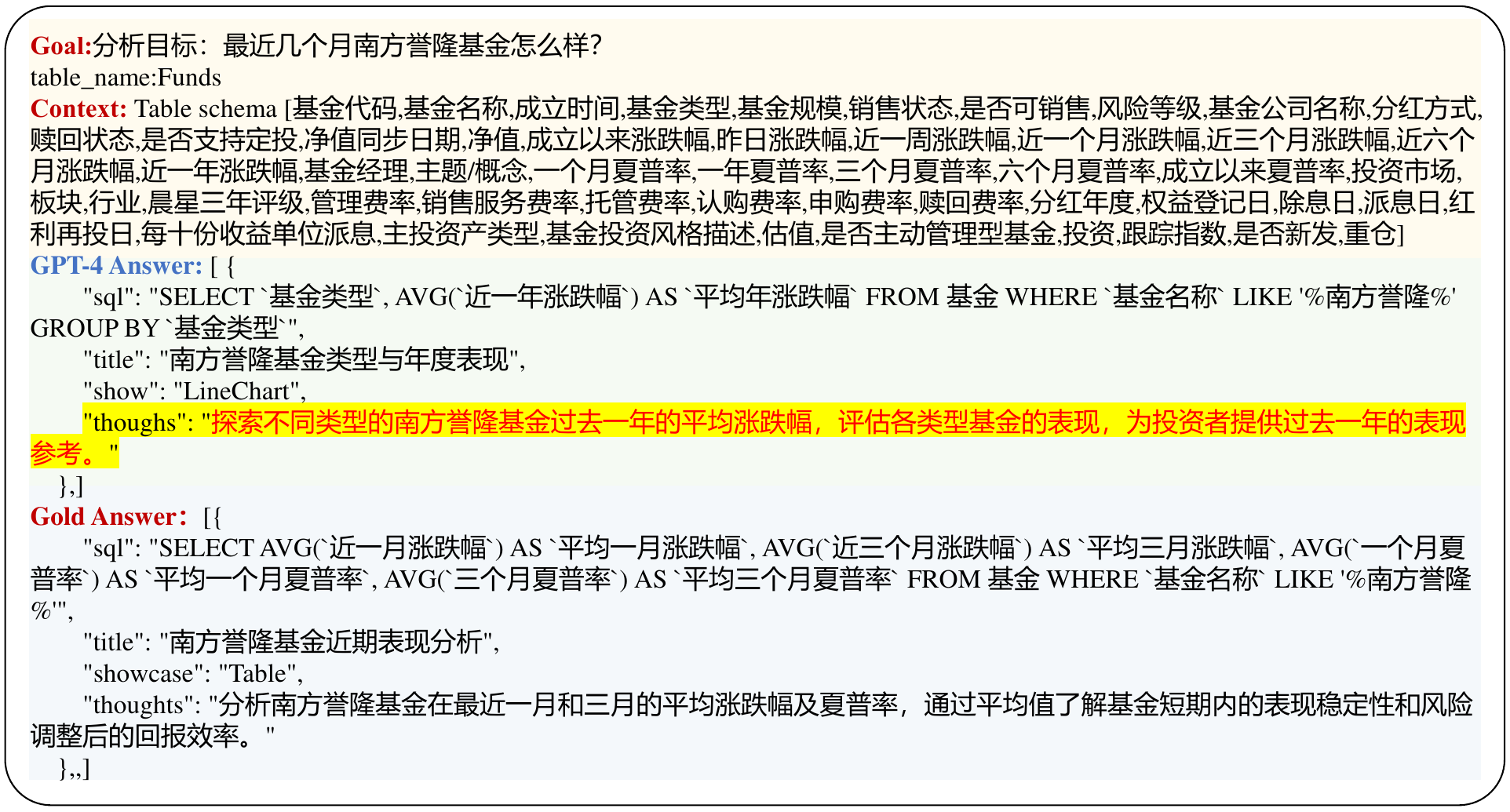}
\caption{
Case study on the NL2ViSQL task, we highlight large language model \illusion{\textit{analysis error}}.
} 
\label{nl2visqlcasestuy}
\end{figure}

\section{In-depth Analysis}
Given the constraints on content, we have selected representative LLMs for in-depth analysis based on their types and scores.

\textbf{SFT may enhance model performance}. As Table~\ref{scorecompare} demonstrates the open-source models' SFT versions outperform their Base counterparts. Notably, SFT data, collected from general domains, significantly improved model performance in financial data analysis tasks. Across models of equal size and architecture, performance variations suggest the training data's scope and specificity impact downstream tasks. 

\textbf{Financial-specific fine-tuning proves beneficial}. To assess the impact of financial domain knowledge fine-tuning, we compared three LLMs, specifically fine-tuned with financial domain knowledge, against their corresponding base models, as shown in Table~\ref{subtaskscore}. Notably, the XuanYuan models demonstrate continuous score improvements after financial-specific knowledge fine-tuning. A closer examination of the 6 sub-tasks reveals that LLaMA2-13B and 70B perform poorly across all tasks, indicating a lack of pre-training on a large-scale, high-quality financial corpus. Nonetheless, fine-tuning with financial knowledge results in significant improvements. However, the models do not excel in tasks 2-3 and 3-1 post-fine-tuning, suggesting that fine-tuning alone may not suffice for complex financial data analysis tasks.

\begin{table}[t!]
\resizebox{\linewidth}{!}{
\begin{tabular}{@{}cccccc@{}}
\toprule
\multirow{2}{*}{Task Name} & \multirow{2}{*}{Metrics} & \multicolumn{2}{c}{Base LLMs}                                & \multicolumn{2}{c}{Fiancial LLMs}                                   \\ \cmidrule(l){3-6} 
                           &                          & \multicolumn{1}{l}{LLaMA2-13B} & \multicolumn{1}{l}{LLaMA2-70B} & \multicolumn{1}{l}{XuanYuan-13B} & \multicolumn{1}{l}{XuanYuan-70B} \\ \midrule
\multicolumn{1}{c|}{1-1}   & \multicolumn{1}{c|}{Acc} & 0.71                         & 8.60                         & 2.05                           & \textbf{11.23}                 \\ \midrule
\multicolumn{1}{c|}{1-2}   & \multicolumn{1}{c|}{F1}  & 0.28                         & 10.27                        & 14.39                          & \textbf{22.40}                 \\
\multicolumn{1}{c|}{2-1}   & \multicolumn{1}{c|}{F1}  & 0.42                         & 0.43                         & 13.97                          & \textbf{21.33}                 \\ \midrule
\multicolumn{1}{c|}{2-2}   & \multicolumn{1}{c|}{R-L} & 1.74                         & 5.15                         & 25.17                          & \textbf{36.28}                 \\
\multicolumn{1}{c|}{2-3}   & \multicolumn{1}{c|}{R-L} & 0.17                         & 3.98                         & 6.24                           & \textbf{7.40}                  \\ \midrule
\multicolumn{1}{c|}{3-1}   & \multicolumn{1}{c|}{EM}  & 1.69                         & \textbf{14.18}               & 14.08                          & 13.12                          \\ \bottomrule
\end{tabular}
}
\caption{
Comparison between different parameter Financial specific LLMs and their base models.} \label{subtaskscore}
\end{table}

\textbf{Most LLMs lack the capability for financial reasoning ablitiy and technical skil}.
As shown in Figure~\ref{Histogram}, we selected five models—GPT-4, GLM4, XuanYuan-70B, Yi-34B, and Internlm-chat-7B—covering a variety of types and model parameters. We display the average scores and variance of these five models across three evaluated dimensions. It is apparent that GPT-4 exhibits a comprehensive advantage in all three categories, particularly in \textit{Foundational Ability} and \textit{Reasoning Ability}, with scores of 41.03 and 32.47, respectively, significantly higher than the other models. This may indicate GPT-4's strong capability in handling financial data analysis regarding foundational and reasoning abilities. Currently, the capabilities of open-source models are generally poor, and even their performance in foundational ability is not ideal. 
Most models, including GPT-4 and GLM-4, show a significant decline in performance on the \textit{Technical Skill} dimension, indicating a lack of data thinking and analytical abilities. 

\textbf{Case Study}. In Figure~\ref{nl2visqlcasestuy}, displaying incorrect analytical results, we noted that GPT-4 lacks essential financial knowledge, failing to properly understand financial reasoning and analysis methods. It mistakenly identifies fund names as financial terminology, highlighting that mastering technical skills is a significant challenge for LLMs.

\section{Conclusion}
In this work, we introduced \texttt{FinDABench}, a benchmark designed to evaluate the capabilities of LLMs in financial data analysis, comprising six tasks across three cognitive dimensions. We conducted a comprehensive examination of 41 LLMs, assessing their performance. The results reveal that current LLMs generally struggle to deliver meaningful data analysis, with poor scores across most tasks. FinDABench is a valuable resource for future research and development financial data analysis.

\section*{Limitations}
\textbf{Insufficient Data Coverage}: Although we have developed a financial data analysis evaluation framework encompassing three dimensions, the number of sub-tasks currently included does not fully cover all the challenges present in the financial data analysis landscape. In future work, we plan to collaborate with professional financial institutions to construct a more comprehensive and robust financial evaluation dataset. This enhancement will better gauge the advancements of large models in handling complex financial scenarios.

\textbf{Inadequate Evaluation Metrics}: The evaluation metrics currently in use are those traditionally applied to NLP tasks. These metrics fail to adequately measure the performance of large models on generative tasks such as Fin-report2Markdown and NL2ViSQL, nor do they reflect the financial data analysis thinking inherent to large models. In the future, we intend to design more appropriate evaluation metrics based on the real-world objectives of financial data analysis, thereby providing a truer reflection of the models' capabilities.

\bibliography{anthology,custom}
\bibliographystyle{acl_natbib}

\newpage
\appendix
\begin{appendix}

\onecolumn
\section{English Version Data examples}
\label{sec:enversiondataexamples}

\begin{figure*}[ht!]
\centering
\includegraphics[width=1\textwidth]{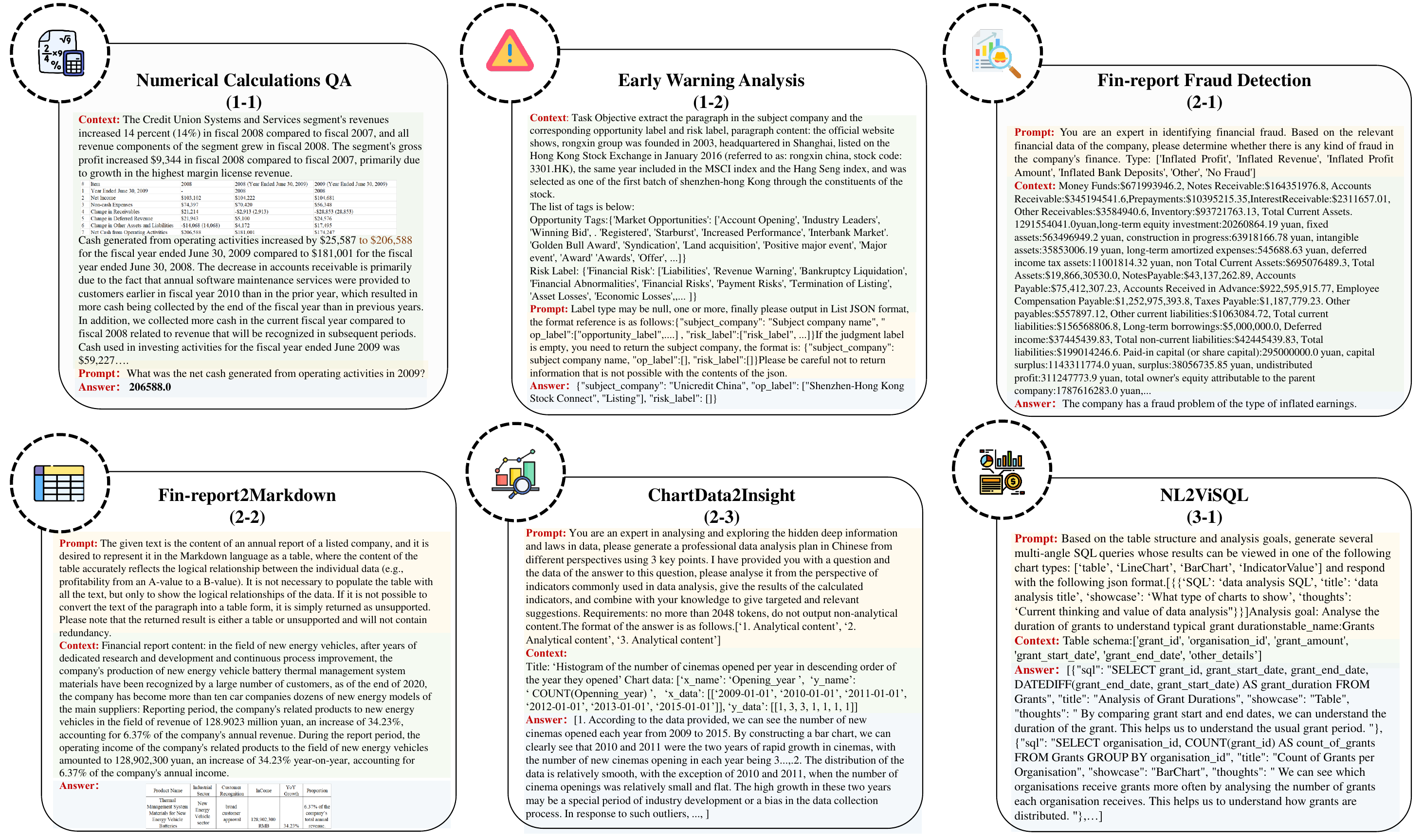}
\caption{
Data examples for the six sub-tasks of \texttt{FinDABench} in English.
} \label{en_datasample}
\end{figure*}

\section{More Details of \texttt{FinDABench}}
\subsection{Prompt Template}
\subsubsection{Numerical Calculation QA Translation Prompt}
\label{sec:numericalcilationqa}

\begin{figure*}[ht!]
\centering
\includegraphics[width=1\textwidth]{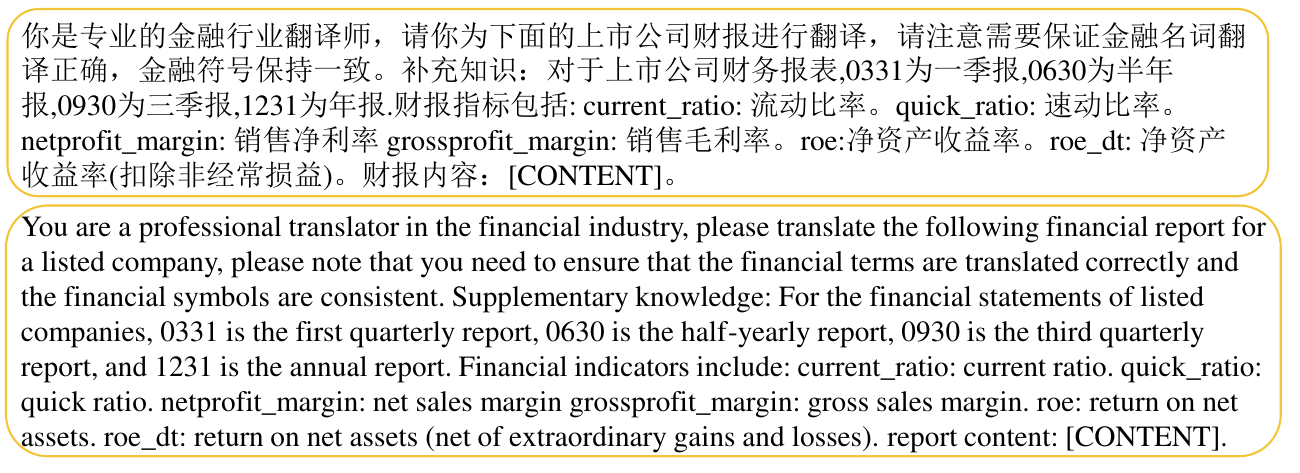}
\caption{
The prompt for translating financial texts into English is displayed above, with the translated version below.
} \label{numerical_calculation_prompt}
\end{figure*}

\subsubsection{Fin-report2Markdown Convert Prompt}
\label{sec:finreport2markdown}

\begin{figure*}[ht!]
\centering
\includegraphics[width=1\textwidth]{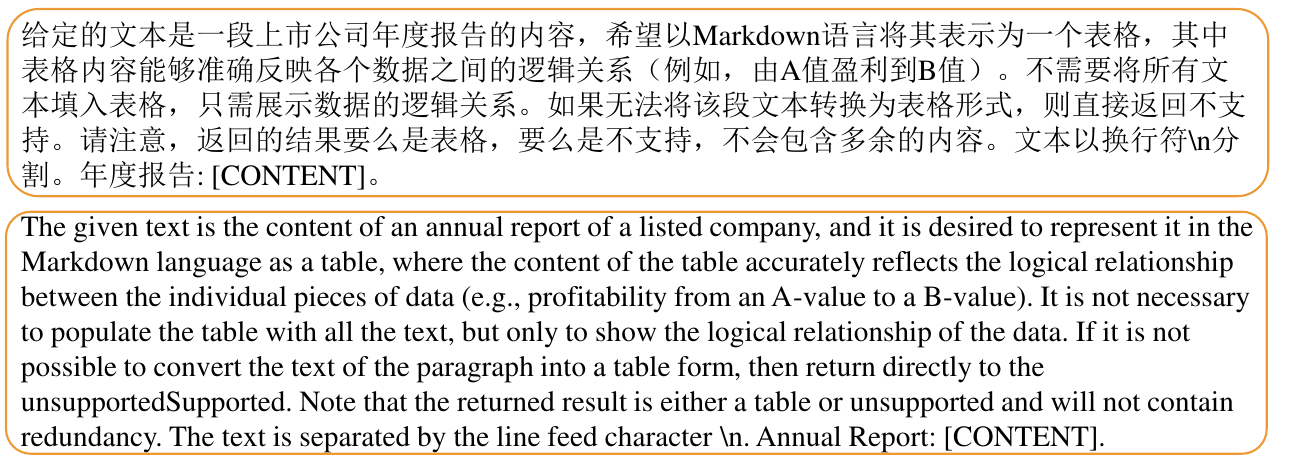}
\caption{
The prompt used for extracting structured Markdown data from an annual report is shown above, with the translated English version presented below.
} \label{finreport2markdownprompt}
\end{figure*}

\subsubsection{ChartData Understanding Prompt}
\label{sec:chartdatainsight}

\begin{figure*}[ht!]
\centering
\includegraphics[width=1\textwidth]{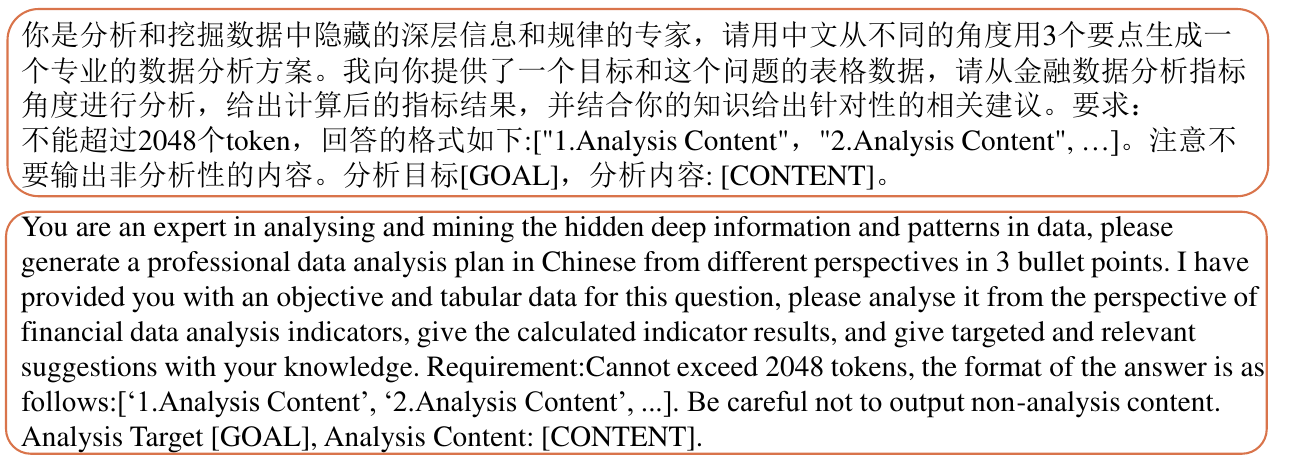}
\caption{
The prompt used for generating analytical insights from chart data is displayed above, with the translated English version provided below.
} \label{chartdatainsightprompt}
\end{figure*}

\subsubsection{NL2ViSQL Prompt}
\label{sec:nl2visql}

\begin{figure*}[ht!]
\centering
\includegraphics[width=1\textwidth]{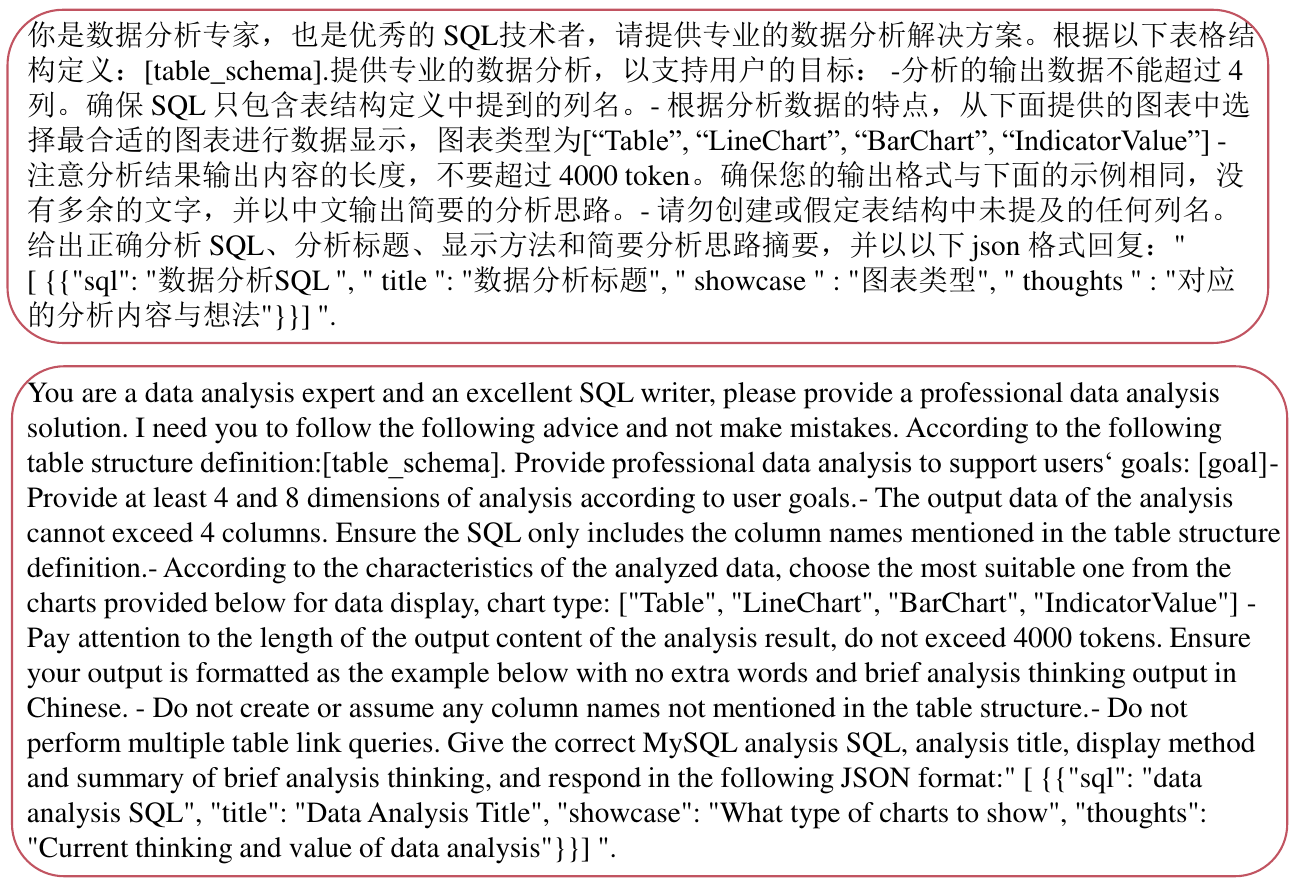}
\caption{
The prompt used for generating multi-perspective SQL based on objectives and table structure is shown above, with the translated English version provided below.
} \label{nl2visqltprompt}
\end{figure*}

\subsection{Early Warning Analysis Label System}
\label{sec:earlywarning}
\begin{figure*}[ht!]
\centering
\includegraphics[width=1\textwidth]{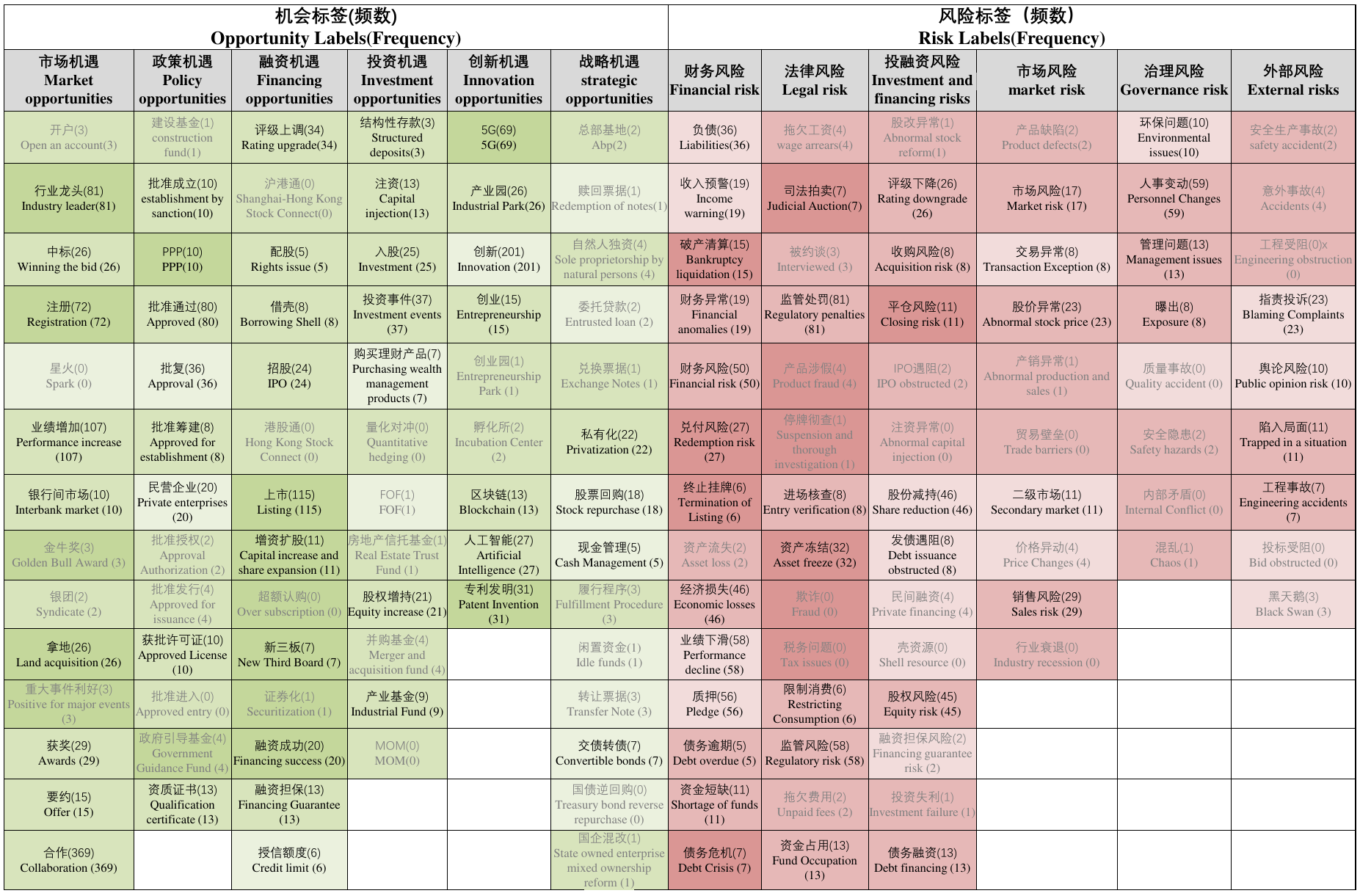}
\caption{The detailed tagging architecture is divided into two main categories: opportunity tags and risk tags. From a financial perspective, it covers sentiment tags throughout the entire lifecycle of a company.
} \label{earlywarning}
\end{figure*}

\section{LLM Test}
\subsection{Large Language Model Test List}
\label{sec:appendixllm}
\begin{table*}[h]
\centering
\resizebox{0.89\textwidth}{!}{
\begin{tabular}{c|c|ccccc}
\hline
Type                                               & Model           & Parameters & Instruction                & RL                      & Access  & BaseModel\\ \hline
\multicolumn{1}{c|}{\multirow{7}{*}{English LLMs}} & GPT-4-0613           & ---        & \ding{51} & \ding{51} & API & ---     \\
\multicolumn{1}{c|}{}                              & GPT-3.5-turbo-0613   & ---        & \ding{51} & \ding{51} & API & ---    \\

\multicolumn{1}{c|}{}                              & LLaMA2-Base     & 7/13/70B   & \ding{51} & \ding{55} & Weights & --- \\
\multicolumn{1}{c|}{}                              & LLaMA2-Chat     & 7/13/70B   & \ding{51} & \ding{51} & Weights & LLaMA2-7/13/70B \\
\multicolumn{1}{c|}{}                              & Vicuna-v1.5     & 7B         & \ding{51}
 & \ding{55} & Weights & LLaMA2-7B \\
\multicolumn{1}{c|}{}                              & Alpaca-v1.0     & 7B         & \ding{51} & \ding{55} & Weights & LLaMA-7B \\
\multicolumn{1}{c|}{}                              & WizardLM        & 7B         & \ding{51} & \ding{55} & Weights & LLaMA-7B\\
\multicolumn{1}{c|}{}                              & Phi        & 2B         & \ding{51} & \ding{55} & Weights & ---\\ \hline
\multirow{19}{*}{Chinese LLMs}                     & \begin{CJK*}{UTF8}{gbsn}通义千问\end{CJK*}(Qwen-turbo)   & ---        & \ding{51} & \ding{51} & API & ---    \\
                                                   & \begin{CJK*}{UTF8}{gbsn}文心一言\end{CJK*}(ERNIEv4.0)   & ---        & \ding{51} & \ding{51} & API & ---    \\
                                                   & \begin{CJK*}{UTF8}{gbsn}
智谱清言\end{CJK*}(GLM-4)   & ---        & \ding{51} & \ding{51} & API & ---    \\
                                                   & Yi-Base & 6B/34B         & \ding{51} & \ding{55} & Weights & --- \\
                                                   & Yi-Chat & 6B/34B         & \ding{51} & \ding{55} & Weights & Yi-6B/34B\\
                                                   & InternLM-Base   & 7B/20B  & \ding{51} & \ding{55} & Weights & --- \\
                                                   & InternLM-Chat   & 7B/20B  & \ding{51} & \ding{55} & Weights &InternLM-7B \\
                                                   & Qwen-Base  & 7B/14B     & \ding{51} & \ding{55} & Weights & ---\\
                                                   & Qwen-Chat  & 1.8B/7B/14B     & \ding{51} & \ding{55} & Weights & Qwen-1.8/7/14B\\
                                                    & Baichuan2-Base  & 7B/13B & \ding{51} & \ding{55} & Weights & ---\\
                                                   & Baichuan2-Chat  & 7B/13B   & \ding{51} & \ding{55} & Weights & Baichuan2-7/13B\\
                                                   & TigerBot-Base        & 7B         & \ding{51} & \ding{55} & Weights & --- \\
                                                   & TigerBot-Chat   & 7B & \ding{51} & \ding{55} & Weights & TigerBot-7B \\
                                                   & Chinese-Alpace2 & 7B         & \ding{51} & \ding{55} & Weights & LLaMA2-7B\\
                                                   & ChatGLM2   & 6B         & \ding{51} & \ding{55} & Weights & ChatGLM-6B \\
                                                   & ChatGLM3-Base   & 6B     & \ding{51} & \ding{55} & Weights & --- \\ 
                                                   & ChatGLM3   & 6B         & \ding{51} & \ding{55} & Weights & ChatGLM3-6B-Base \\
                                                   & MiniCPM   & 2B         & \ding{51} & \ding{55} & Weights &  --- \\
                                                   \hline
\multirow{1}{*}{Financial LLMs}                           & XuanYuan-Chat        & 13/70B        & \ding{51} & \ding{55} & Weights & LLaMA2-13/70B \\ \hline
\end{tabular}
}
\caption{LLMs tested on \texttt{FinDABench}. We classify these models by their main training corpora.} \label{LLMTest}
\end{table*}

\end{appendix}

\end{document}